\documentclass[lettersize,journal]{IEEEtran}

\usepackage{cuted}
\usepackage{graphicx}
\usepackage{cite}
\usepackage{picinpar}
\usepackage{amsmath}
\usepackage{url}
\usepackage[utf8]{inputenc}
\usepackage{colortbl}
\usepackage{soul}
\usepackage{multirow}
\usepackage{pifont}
\usepackage{color}
\usepackage[dvipsnames]{xcolor}
\usepackage{alltt}
\usepackage{hyperref}
\usepackage{enumerate}
\usepackage{siunitx}
\usepackage{breakurl}
\usepackage{epstopdf}
\usepackage{pbox}

\usepackage[ruled,linesnumbered, lined, noend]{algorithm2e}
\usepackage{booktabs}
 
\usepackage{bm}
\usepackage{amssymb}
\usepackage{romannum}
\usepackage{bbm}
\usepackage[caption=false,font=footnotesize]{subfig}

\usepackage{booktabs} 
\usepackage{adjustbox}
\usepackage{siunitx} 
\usepackage{hyperref}
\newcommand{\email}[1]{\href{mailto:#1}{\nolinkurl{#1}}}

\usepackage[normalem]{ulem}

\hypersetup{hidelinks} 
\usepackage[switch]{lineno}

\newcommand{\blue}[1]{{\color{blue} #1}}

\usepackage{siunitx}

\usepackage[multiple]{footmisc}

\begin{document}
\pagenumbering{arabic}

\title{
Failure-Aware Bimanual Teleoperation via Conservative Value Guided Assistance
}  

\author{
Peng Zhou$^{*}$,~\IEEEmembership{Member,~IEEE},
Zhongxuan Li$^{*}$,
Jinsong Wu,
Jiaming Qi,
Jun Hu,
\\
David Navarro-Alarcon,~\IEEEmembership{Senior Member,~IEEE},
Jia Pan,~\IEEEmembership{Senior Member,~IEEE},
Lihua Xie,~\IEEEmembership{Fellow,~IEEE}, \\
Shiyao Zhang$^{\dag}$,~\IEEEmembership{Senior Member,~IEEE}
and Zeqing Zhang$^{\dag}$,~\IEEEmembership{Member,~IEEE}

\thanks{
This work was supported by the National Natural Science Foundation of China (NSFC) under Grant No. 62403211, in part by Youth S\&T Talent Support Programme of Guangdong Provincial Association for Science and Technology (GDSTA) under Grant No. SKXRC2025092, and in part by Guangdong Regional Joint Fund for Basic and Applied Basic Research Fund (No. 2024A1515110203).
\emph{(Corresponding author: Shiyao Zhang, Zeqing Zhang)}
}

\thanks{$^{*}$ indicates that the authors contributed equally to this work.} 

\thanks{P. Zhou, J. Hu and S. Zhang are with the School of Advanced Engineering, Great Bay University, Guangdong, China
(e-mail: \email{pzhou@gbu.edu.cn}, \email{hujun@gbu.edu.cn}, \email{zhangshiyao@gbu.edu.cn}).}

\thanks{Z. Li and J. Pan are with the School of Computing and Data Science, The University of Hong Kong, Hong Kong
(e-mail: \email{zxli01@connect.hku.hk}, \email{jpan@cs.hku.hk}).}

\thanks{J. Wu and D. Navarro-Alarcon are with the Department of Mechanical Engineering, The Hong Kong Polytechnic University, Kowloon, Hong Kong (e-mail: \email{jinsong.wu@connect.polyu.hk}, \email{dnavar@polyu.edu.hk}).}

\thanks{J. Qi is with the College of Mechanical and Electrical Engineering, Northeast Forestry University, Heilongjiang, China
(e-mail: \email{jiamingqi@nefu.edu.cn}).}

\thanks{L. Xie and Z. Zhang are with the School of Electrical and Electronic Engineering, Nanyang Technological University, Singapore
(e-mail: \email{elhxie@ntu.edu.sg}, \email{zeqing.zhang@ntu.edu.sg}).}

}

\maketitle

\begin{abstract}
Teleoperation of high-precision manipulation is constrained by tight success tolerances and complex contact dynamics, which make impending failures difficult for human operators to anticipate under partial observability. This paper proposes a value-guided, failure-aware framework for bimanual teleoperation that provides compliant haptic assistance while preserving continuous human authority. The framework is trained entirely from heterogeneous offline teleoperation data containing both successful and failed executions. Task feasibility is modeled as a conservative success score learned via Conservative Value Learning, yielding a risk-sensitive estimate that remains reliable under distribution shift. During online operation, the learned success score regulates the level of assistance, while a learned actor provides a corrective motion direction. Both are integrated through a joint-space impedance interface on the master side, yielding continuous guidance that steers the operator away from failure-prone actions without overriding intent. Experimental results on contact-rich manipulation tasks demonstrate improved task success rates and reduced operator workload compared to conventional teleoperation and shared-autonomy baselines, indicating that conservative value learning provides an effective mechanism for embedding failure awareness into bilateral teleoperation. Experimental videos are available at
\blue{\url{https://www.youtube.com/watch?v=XDTsvzEkDRE}}
\end{abstract}


\IEEEpeerreviewmaketitle

\begin{figure}[h]
\centering
\includegraphics[width=0.99\linewidth]{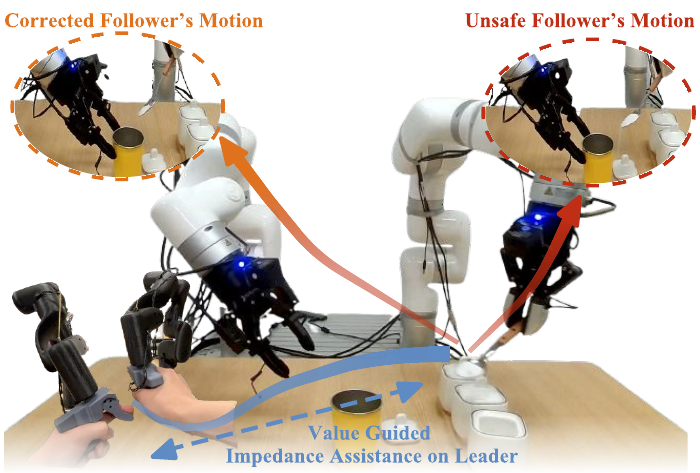}
\caption{Our failure-aware teleoperation framework leverages conservative success score learning to trigger corrective torque on the leader arm, guiding the operator away from out-of-distribution (OOD) behaviors.}
\label{fig:intro-1}
\vspace{-0.5cm}
\end{figure}

\section{Introduction}
Teleoperation serves as a critical infrastructure for both collecting large-scale, high-fidelity human demonstrations and executing robotic manipulation in hazardous or inaccessible environments. As data-driven approaches, such as imitation learning and offline reinforcement learning—increasingly rely on diverse and precise teleoperation data to acquire dexterous manipulation skills, the quality and continuity of these trajectories play a central role in determining policy performance~\cite{qin2023anyteleop}. At the same time, teleoperation remains the only viable solution for safety-critical tasks in which direct human intervention is impossible. Together, these demands require fine-grained control over delicate physical interactions, despite persistent challenges such as partial observability, complex contact dynamics, and communication latency. Consequently, there is an urgent need for advanced teleoperation systems capable of anticipating and mitigating such irreversible failures before they occur.

Extensive research \cite{liu2025factr,wuphilipp2024gello} has sought to improve teleoperation performance through enhanced communication stability and high-fidelity force and position feedback to achieve bilateral transparency. While these advances improve signal transmission accuracy, the information is typically treated as passive feedback and rarely exploited to reason about task context or failure risk, leaving most systems reliant on human operators to compensate for uncertainties. In contact-rich manipulation, effective teleoperation requires not only faithful signal transmission but also motion-level intelligence on the robot side to interpret commands within the local physical context. The absence of such capability prevents small execution deviations from being corrected proactively, allowing them to accumulate into irreversible failures. Moreover, communication latency and partial observability create a persistent perception gap, making sole reliance on human compensation insufficient for robust performance.

Recent learning-based paradigms \cite{an2025dexterous} demonstrate the potential of equipping robots with dexterous manipulation skills learned from simulation or large-scale data, while allowing humans to provide high-level task guidance during execution. In these frameworks, the robot autonomously generates low-level motions based on learned policies, and the human influences behavior primarily through coarse command modulation or task-level intent. However, the learned policy operates largely independently of the operator’s continuous control loop, providing limited feedback to the human about local execution quality or emerging failure risk. As a result, discrepancies between the robot’s autonomous actions and the operator’s expectations can accumulate during teleoperation, leading to reduced coordination and increased sensitivity to contact-induced deviations in fine-grained manipulation.

In parallel, shared autonomy frameworks \cite{dragan2013policy} have introduced artificial intelligence to assist at the task level, for example, through intent inference or strategic planning. These methods, however, typically operate at an abstract level and lack the control resolution required to modulate fine-grained motions during delicate, contact-rich interactions. As a result, existing teleoperation and shared autonomy systems tend to respond to failures only after they become apparent , leaving the human–robot team vulnerable to irreversible outcomes that could have been mitigated through earlier, motion-level intervention.

To address these challenges, we propose a framework that explicitly decouples failure awareness from online action execution. The core idea is to infer, from offline teleoperation data, a conservative estimate of task feasibility that evaluates whether a given state–action pair is likely to remain within a success-consistent region of the task. Rather than directly altering or overriding human commands, this feasibility estimate is used to regulate interaction in a minimally invasive manner. 

Our main contributions are summarized as follows:
\begin{itemize}
    \item We formulate a \emph{failure-aware success score} learned from heterogeneous offline teleoperation data, which provides a conservative estimate of task feasibility and remains reliable under operator-induced distribution shift.
    \item We develop a \emph{success-guided assistance paradigm} that leverages the learned success score to regulate interaction strength while preserving continuous human control and avoiding discrete autonomy switching.
    \item We realize this assistance through a \emph{bilateral teleoperation interface} in which the robot’s corrective intent is rendered as compliant, physically grounded haptic feedback, biasing human commands toward success-consistent execution without overriding operator authority.
\end{itemize}

Through this formulation, the robot complements human control with predictive failure awareness, enabling safer, more reliable teleoperation in contact-rich domains and supporting scalable data collection for learning-based manipulation.

\section{Related Work}

\subsection{Bimanual Telemanipulation Systems}
Bimanual telemanipulation systems enable dexterous manipulation by leveraging direct human control for high–DoF tasks and are widely used in contact-rich scenarios where fully autonomous policies remain unreliable. A representative example is ALOHA \cite{fu2024mobile}, a low-cost, open-source bimanual teleoperation platform that provides coordinated dual-arm control through a leader–follower interface. Several works focus on improving the usability and feedback structure of teleoperation interfaces. GELLO \cite{wuphilipp2024gello} introduces a kinematically matched operator interface to improve control intuitiveness and demonstration quality, while FACTR \cite{liu2025factr} reflects follower-side motor currents to convey contact information but lacks impedance modulation or motion-level stabilization on the master side, resulting in limited feedback under rapidly changing contact conditions. In parallel, recent learning-based approaches explore teleoperation as a high-level intent interface for pretrained controllers. For example, DexterityGen~\cite{yin2025dexteritygen} leverages large-scale simulation and reinforcement learning to acquire dexterous skills, with human inputs providing coarse guidance. In these systems, however, fine-grained manipulation is largely executed autonomously by the robot, with limited continuous human--robot coordination or bidirectional feedback during task execution. Overall, existing bimanual telemanipulation systems highlight the value of human control for dexterous manipulation and data collection, while exposing limitations in the integration of robot-side intelligence and physically grounded feedback within the teleoperation loop.

\subsection{Failure-Aware Shared Autonomy and Teleoperation}
Shared autonomy (SA)\cite{dragan2013policy, lasota2017survey} has been widely studied to reduce human cognitive load by blending operator inputs with autonomous decision-making. Early SA frameworks often modeled the problem as a Partially Observable Markov Decision Process (POMDP)\cite{cassandra1998survey}, enabling intent inference and goal-oriented assistance under perceptual uncertainty~\cite{javdani2018shared}. While effective for high-level reasoning, these approaches typically rely on simplified task abstractions and provide limited support for fine-grained motion regulation. With increasing safety demands in contact-rich manipulation, recent work has emphasized safety-critical control filters \cite{aigner1997human, bansal2017hamilton} to constrain system behavior within predefined safe sets without fully overriding human control~\cite{mower2021skill, muelling2017autonomy}. Although these methods offer formal safety guarantees, they are largely reactive and require explicit constraint modeling, which is difficult for complex contact dynamics. To improve adaptability, subsequent research incorporated human interventions into imitation learning pipelines. Approaches such as Human–Agent Joint Learning~\cite{liu2025robot} and its variants~\cite{luo2025human, schaff2020residual, reddy2018shared} allow experts to correct robot behavior online, mitigating covariate shift, but primarily address observed errors rather than anticipating failures. More recently, foundation-model-based approaches have explored using vision–language models (VLMs) to provide task-level semantic guidance~\cite{liu2025casper}. However, these methods operate at coarse temporal and spatial resolutions and do not capture local contact dynamics. As a result, despite substantial progress, most existing shared autonomy approaches remain reactive or focused on task-level guidance, lacking motion-level understanding and predictive feedback for impending, often irreversible failures such as jamming or misalignment.

\subsection{Offline Reinforcement Learning for Policy Improvement}
Offline reinforcement learning (RL)\cite{levine2020offline, watkins1992q} enables policy improvement from fixed datasets without online exploration. Methods such as Implicit Q-Learning (IQL)\cite{kostrikov2021offline, ziebart2008maximum} improve offline stability by avoiding explicit maximization over unseen actions, while Conservative Q-Learning (CQL)~\cite{kumar2020conservative} further addresses distribution shift by suppressing Q-values for out-of-distribution actions, yielding more reliable value estimates in contact-rich settings. Despite progress in offline RL for autonomous manipulation, including large-scale efforts such as $\pi_{0.6}^{*}$\cite{pi0_2025} and RL100\cite{lei2025rl}, most existing methods are designed for trajectory-level optimization and autonomous execution~\cite{xu2022look}. Consequently, they provide limited support for shared autonomy and interactive teleoperation, as learned policies rarely expose feasibility or uncertainty information in a form that can be intuitively communicated to human operators~\cite{schwarting2017parallel}. This highlights a key limitation of prior work: although offline RL captures task success structure from data, it is seldom integrated into human–robot control loops in a manner that preserves continuous human authority \cite{zhang2025collaborative}. Addressing this gap requires decoupling failure awareness learned from offline data from direct action execution, and instead conveying it through interpretable, physically grounded guidance signals that support human decision making.

\section{Problem Formulation}
We study bilateral telemanipulation with a human-operated master device and a slave robot interacting with an environment. The human issues commands while the robot executes them under partial observability, contact uncertainty, and delay. Our objective is to provide \emph{failure-aware assistance} that reduces irreversible failures while preserving human authority.

\subsection{Telemanipulation as a Failure-Aware POMDP}
We model the interaction process as a \emph{failure-aware }Partially Observable Markov Decision Process (FA-POMDP).
\begin{equation}
\mathcal{M}=\langle \mathcal{X},\mathcal{A},\mathcal{S},\Omega,\mathcal{P}, \mathcal{R}\rangle,
\end{equation} where $\mathcal{X}$ denotes the underlying physical dynamics of the robotic manipulation system, represented by its state space, including object poses, contact modes, and frictional parameters. $\mathcal{A}$ indicates the velocity commands $a_{tele}$ on slave robot. These observations are generated from the underlying system state through a stochastic transition model
\begin{equation}
x_{t+1} \sim \mathcal{P}(x_{t+1} \mid x_t, a_t),
\end{equation}
which captures hybrid contact dynamics and object–robot interactions. 

Given the partially observable nature of teleoperation, at each time step the human operator perceives a state $s_t \in \mathcal{S}$ composed of leader–follower joint velocities $q_t$ and visual observations $o_t$. The reward function $R: \mathcal{S} \times \mathcal{A} \rightarrow \mathbb{R}$ provides a failure-aware supervision signal that characterizes the task outcome. Instead of relying on manual reward engineering, we employ a binary signal $r_t \in \{+1, -1\}$ broadcasted across the trajectory based on its terminal status. Specifically, $r_t = +1$ indicates the system remains within a success-consistent region, while $r_t = -1$ denotes a transition toward an irreversible failure or a state requiring human intervention. 

\subsection{The Overall Problem Objective}

The objective of this work is to learn a conservative feasibility estimator $Q$ from heterogeneous offline teleoperation data, and to derive an assisted policy $\pi_{\text{assist}}$ that proactively mitigates failure risk while preserving continuous human control authority.

This objective is formulated as a constrained optimization problem:
\begin{equation}
\min_{Q, \lambda} \quad 
\mathbb{E}_{\pi_{\text{assist}}} \!\left[ 
\sum_{t=0}^{\infty} \gamma_{\text{risk}}^t \, \mathbb{I}[x_t \in \mathcal{F}_{\text{failure}}] 
\right],
\end{equation}
subject to a minimal-invasiveness constraint,
\begin{equation}
\mathbb{E}_{s_t \sim d^{\pi_{\text{assist}}}} 
\!\left[ \left| a_t - a_t^{\text{tele}} \right| \right] 
\le \epsilon,
\end{equation}
and a gated transparency condition,
\begin{equation}
\lambda_t = 0 \quad \text{if} \quad Q(s_t, a_t^{\text{tele}}) \ge \tau.
\end{equation}

The assisted policy $\pi_{\text{assist}}$ is induced by a command blending mechanism. In this formulation, $\lambda_t \in [0,1]$ denotes a time-varying \emph{intervention weight} that modulates the influence of robot assistance in the executed command. Rather than being a free control variable, $\lambda_t$ is a value-gated quantity induced by the learned feasibility score $Q$, reflecting the estimated near-future failure risk under the operator’s current command.

Intuitively, when the operator command is assessed as safe (high $Q$ and low predicted risk), $\lambda_t \approx 0$ and the system behaves transparently. As the predicted risk increases, $\lambda_t$ grows smoothly, activating success-guided assistance that biases execution toward safer actions without overriding human intent.

By introducing a dedicated failure set $\mathcal{F}_{\text{failure}}$, we explicitly model the cumulative nature of telemanipulation failures, which typically arise from multi-step state evolution under hidden contact dynamics rather than isolated erroneous actions. Entering $\mathcal{F}_{\text{failure}}$ corresponds to an irreversible terminal event that triggers a system reset by the human operator. Our goal is therefore to identify the decision boundary between the safety set $\mathcal{F}_{\text{safe}}$ and the failure set $\mathcal{F}_{\text{failure}}$, enabling early detection of imminent risks and reducing the probability of system resets.

The proposed framework consists of three key components:  
(1) a \textit{Conservative Success Score Estimator} learned from offline teleoperation data to characterize feasibility under distribution shift;  
(2) a \textit{Success-Guided Actor} that provides online corrective motion directions aligned with safer regions of the learned feasibility landscape; and  
(3) a \textit{Value-Guided Impedance Assistance Module} that operates transparently on the leader arm, delivering physically grounded haptic feedback to steer the operator away from failure-prone manipulations while preserving continuous human control.

\section{Methodology}


\begin{figure}[htbp]
  \centering
  \includegraphics[width=0.99\linewidth]{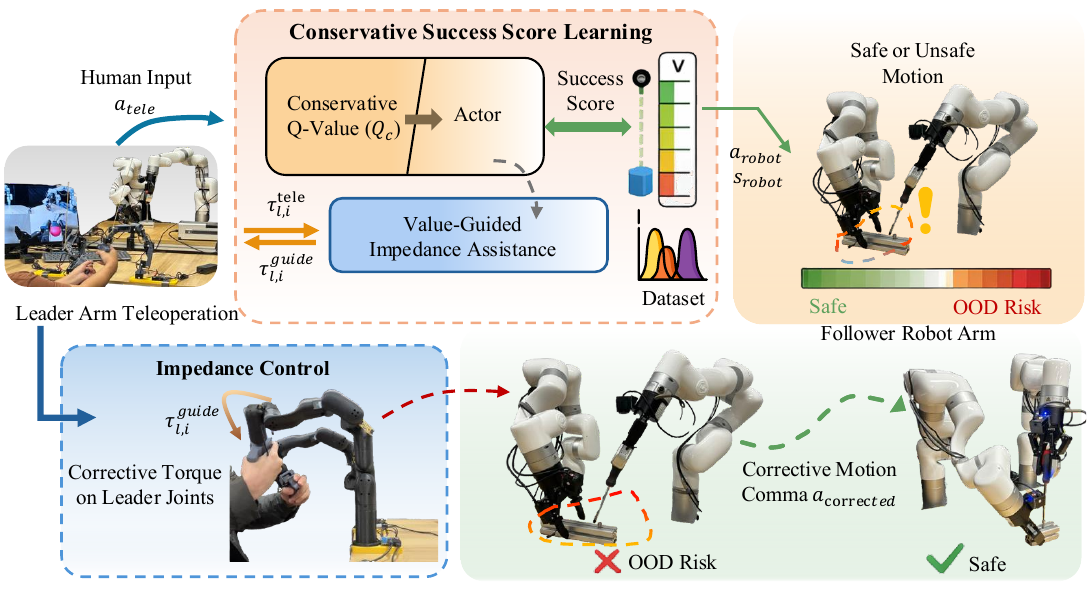}
  \caption{\textbf{Overview of the proposed value-guided shared autonomy framework.} Upper part: Offline learning from heterogeneous teleoperation data with binary success and failure labels to train a conservative success score (critic). Bottom part: Online deployment, where the success score determines when intervention is required, and a learned actor provides a corrective guidance action. Both are integrated within a shared autonomy framework to guide the operator away from out-of-distribution and failure-prone actions while preserving continuous human control.
}
  \label{fig:method-1}
\end{figure}

This section presents the overall system architecture and learning framework for success-guided teleoperation assistance. We first introduce the offline learning pipeline for training a conservative feasibility estimator from heterogeneous teleoperation data. We then describe the real-time inference and decision-making process that integrates the learned success signal into a bilateral teleoperation system, enabling risk-aware command modulation and proactive intervention. Finally, we detail the control interface and coupling mechanisms that ensure seamless, transparent, and physically grounded human--robot interaction.

\subsection{Offline Teleoperation Dataset Collection}\label{method:collection}

All experiments use an offline dataset collected exclusively through bilateral teleoperation, without any autonomous assistance. Data are gathered from both expert operators and novice users to capture diverse execution styles, corrective behaviors, and representative failure patterns. Operators control the slave robot via a master device in joint velocity space and receive visual feedback from a multi-view RGB camera setup, including external front, left, and right cameras as well as wrist-mounted cameras on both end-effectors. This configuration captures both global scene context and fine-grained contact interactions.

We assume access to heterogeneous offline teleoperation data
\begin{equation}
\mathcal{D} = \mathcal{D}_{\text{succ}} \cup \mathcal{D}_{\text{fail}},
\end{equation}
where $\mathcal{D}_{\text{succ}}$ contains successful executions and $\mathcal{D}_{\text{fail}}$ contains trajectories that lead to irreversible failures or near-failure events. 

Each teleoperation episode is limited to 30 seconds and terminates upon task completion, explicit failure indication, or timeout, yielding temporally compact trajectories suitable for offline learning. Each trajectory is represented as
\begin{equation}
\tau = \{(s_t, a_t^{\text{tele}}, s_{t+1})\}_{t=0}^{T},
\end{equation}
where $s_t$ denotes the observed system state and $a_t^{\text{tele}}$ is the joint-velocity command issued by the human operator. No corrective or autonomous actions are applied during data collection.

Rather than encoding task efficiency or incremental progress, we adopt a failure-aware supervision signal based on trajectory-level outcomes. After execution, each trajectory is labeled as either successful or failed, and the corresponding binary label is assigned uniformly to all time steps:
\begin{equation}
r_t =
\begin{cases}
+1, & \tau \in \mathcal{T}_{\text{success}}, \\
-1, & \tau \in \mathcal{T}_{\text{failure}}.
\end{cases}
\end{equation}

We denote the offline dataset as
\begin{equation}
\mathcal{D}=\{\tau^{(n)}\}_{n=1}^{N}, \qquad 
\tau^{(n)}=\{(s_t^{(n)}, a_t^{\text{tele},(n)}, s_{t+1}^{(n)})\}_{t=0}^{T_n-1},
\label{eq:dataset_def}
\end{equation}
where $s_t \in \mathcal{S}$ is the observed state and $a_t^{\text{tele}} \in \mathcal{A}$ is the human-issued joint-velocity command. The resulting dataset serves as the sole source of supervision for learning the task-relevant value function in the following section.

\subsection{Conservative Success Score Learning}
we learn a conservative success score $Q(s,a)$ that serves as a near-future feasibility estimator for teleoperation commands. The score is defined as the expected cumulative discounted reward:
\begin{equation}
Q(s_t, a_t) = \mathbb{E}\!\left[\sum_{i=0}^{\infty} \gamma^i r_{t+i} \,\middle|\, s_t, a_t \right],
\label{eq:Q-value}
\end{equation}
where the per-step supervision signal $r_t \in \{+1, -1\}$ is assigned according to the terminal outcome of the trajectory.

Low values of $Q(s_t, a_t^{\text{tele}})$ indicate that the current state--action pair either correlates with failed executions in the offline dataset or deviates from the support of successful demonstrations, signalling elevated near-future failure risk. Accordingly, we define the feasibility of a teleoperation command via the threshold-based rule:
\begin{equation}
(s_t, a_t^{\text{tele}}) \in
\begin{cases}
\mathcal{F}_{\text{feasible}}, & \text{if } Q(s_t, a_t^{\text{tele}}) \ge \tau, \\
\mathcal{F}_{\text{failure}}, & \text{if } Q(s_t, a_t^{\text{tele}}) < \tau,
\end{cases}
\end{equation}
where $\tau$ is a safety threshold determined from the empirical distribution of successful trajectories. This formulation enables proactive identification of impending failures and supports timely, risk-aware intervention in contact-rich teleoperation scenarios.

Our objective is to instantiate the abstract feasibility indicator $Q(s,a)$ as a learnable, reliable, and conservative success score from offline teleoperation data. The learned score is required to provide a calibrated estimate of near-future failure risk under distribution shift and to support real-time intervention gating during teleoperation.

We adopt Conservative Q-Learning (CQL) as an offline reinforcement learning framework for learning such a feasibility estimator from fixed teleoperation datasets. CQL is specifically designed to mitigate overestimation on out-of-distribution (OOD) actions by enforcing a pessimistic value structure, which is critical for safety-sensitive teleoperation scenarios.

We parameterize the success score as an action-value function
\begin{equation}
Q_{\phi}(s,a): \mathcal{S} \times \mathcal{A} \rightarrow \mathbb{R},
\end{equation}
where $\phi$ denotes the network parameters. Unlike conventional reinforcement learning, where the value function is optimized to maximize expected return, here $Q_{\phi}$ is interpreted as a \emph{feasibility potential} that measures the statistical likelihood that executing action $a$ at state $s$ will remain within a success-consistent region of the task. Lower values indicate proximity to irreversible failure modes, while higher values correspond to actions supported by successful teleoperation data.

The critic is trained by minimizing the CQL objective
\begin{equation}
\begin{aligned}
\mathcal{L}_{\text{critic}}(\phi)
=&\;
\frac{1}{2}
\mathbb{E}_{(s,a,s')\sim\mathcal{D}}
\Big[
\big(
Q_{\phi}(s,a)
-\hat{\mathcal{B}}Q_{\bar{\phi}}(s,a)
\big)^2
\Big]
\\[-0.3em]
+&\;
\alpha\,
\mathbb{E}_{s\sim\mathcal{D}}
\Big[
\log\!\sum_{a}\exp(Q_{\phi}(s,a))
-
\mathbb{E}_{a\sim\mathcal{D}}Q_{\phi}(s,a)
\Big]
\end{aligned}
\label{eq:cql_loss}
\end{equation}
where $\hat{\mathcal{B}} Q_{\bar{\phi}}$ denotes the Bellman backup computed using a target network with parameters $\bar{\phi}$, and $\alpha>0$ controls the strength of conservative regularization. The second term explicitly penalizes high values assigned to actions outside the support of the offline dataset, enforcing
\begin{equation}
Q_{\phi}(s,a_{\text{OOD}}) \le Q_{\phi}(s,a_{\text{data}}),
\end{equation}
and thereby preventing optimistic extrapolation under distribution shift.

To sharpen the learned decision boundary near failure-prone regions, we augment the critic with an auxiliary head that predicts whether an irreversible failure will occur within the next $H$ steps. Given the observation--action pair $(s_t,a_t)$, the auxiliary head outputs a failure probability
\begin{equation}
\hat{p}_t^{(H)} = \sigma\!\left(g_{\eta}(s_t,a_t)\right),
\end{equation}
where $g_{\eta}$ is a lightweight predictor sharing the critic backbone. The corresponding short-horizon failure label is defined as
\begin{equation}
y_t^{(H)} =
\mathbb{I}\!\left[
\bigcup_{k=1}^{H} \{x_{t+k}\in\mathcal{F}_{\text{failure}}\}
\right].
\end{equation}

The auxiliary head is trained using a binary cross-entropy loss,
\begin{equation}
\mathcal{L}_{\text{fail}}(\eta)
=
\mathbb{E}_{(s_t,a_t)\sim\mathcal{D}}
\big[
\mathrm{BCE}\big(y_t^{(H)},\,\hat{p}_t^{(H)}\big)
\big].
\end{equation}
This short-horizon supervision complements trajectory-level outcomes and improves the sensitivity of the conservative success score near the safety--failure boundary, without altering the underlying offline RL objective.

Overall, The critic is trained by minimizing the combined loss
\begin{equation}
\mathcal{L}_{\text{critic}}
=
\mathcal{L}_{\text{CQL}}
+
\lambda_{\text{fail}}\,\mathcal{L}_{\text{fail}},
\end{equation}
where $\mathcal{L}_{\text{CQL}}$ enforces conservative Bellman consistency on the success score and $\mathcal{L}_{\text{fail}}$ provides short-horizon failure supervision, with $\lambda_{\text{fail}}$ balancing the two terms.

After training, the raw critic outputs are normalized into a bounded success indicator
\begin{equation}
\tilde{Q}(s,a)
=
\mathrm{clip}
\!\left(
\frac{Q_{\phi}(s,a)-Q_{\min}}{Q_{\max}-Q_{\min}},
\,0,\,1
\right),
\label{eq:normlalize}
\end{equation}
where $Q_{\min}$ and $Q_{\max}$ are computed from the empirical distribution of $Q_{\phi}$ over the training set. In practice, $\tilde{Q}(s,a)\in[0,1]$ provides a dimensionless feasibility margin that is used for value-based intervention gating and guidance modulation in the teleoperation control loop. The resulting conservative success landscape exhibits high-score basins aligned with successful demonstrations and sharp gradients near failure-prone regions, enabling reliable near-future risk assessment from partial observations.

\subsection{Actor Policy as Success-Guided Assistance}
Given the current observation $s_t$, the actor outputs a robot-suggested command
\begin{equation}
a_t^{\text{assist}} = \pi_{\theta}(s_t), \qquad \pi_{\theta}:\mathcal{S}\rightarrow\mathcal{A}.
\label{eq:actor_def}
\end{equation}
We train $\pi_\theta$ to propose actions with high conservative success score under $Q_\phi$, while remaining smooth and within actuation limits. The actor's objective is formulated as:
\begin{equation}
\mathcal{L}_{\pi}(\theta)
=
\mathbb{E}_{s\sim\mathcal{D}}
\Big[
-\,Q_{\phi}\big(s,\pi_{\theta}(s)\big)
+\|\pi_{\theta}(s)-a^{\text{tele}}\|^2
\Big],
\label{eq:actor_loss}
\end{equation}
where $\beta_{\pi}>0$ regularizes magnitude (and can be extended to include jerk/smoothness penalties). This update steers the policy toward actions that locally reduce predicted failure risk, while preserving closeness to the human operation commands during training.

From a control-theoretic standpoint, the actor can be interpreted as learning a feedback control law that follows the negative gradient of a Lyapunov-like feasibility potential,
\begin{equation}
a^{\text{assist}}(s) \;\approx\; -K\,\nabla_a Q_{\phi}(s,a),
\end{equation}
where $K$ is an implicit gain determined by the network parameterization. This perspective clarifies the role of the actor as a corrective motion generator that biases execution toward safer regions without encoding explicit task goals.

Together, the critic and actor instantiate a learned safety module that couples conservative feasibility estimation with a principled, gradient-based corrective direction, enabling proactive and minimally invasive assistance in contact-rich teleoperation, as described in Algorithm \ref{alg:offline_training}

\LinesNotNumbered
\begin{algorithm}[htbp]
\caption{Offline Training of Conservative Success Score and Guidance Actor}
\label{alg:offline_training}

\KwIn{Teleoperation episodes $\{\tau\}$}
\KwIn{Discount $\gamma$, conservatism weight $\alpha$, anchor weight $\lambda_{\text{anchor}}$, target period $N_{\text{tgt}}$}
\KwIn{(Optional) auxiliary weight $\lambda_{\text{fail}}$}
\KwOut{Critic $Q_\phi(s,a)$ and actor $\pi_\theta(s)$}

\textbf{(1) Dataset construction.}\;
Collect trajectories $\tau=\{(s_t,a_t^{\text{tele}},s_{t+1})\}$ and form dataset $\mathcal{D}$\;
Assign trajectory label $y(\tau)\in\{+1,-1\}$ and set $r_t\leftarrow y(\tau)$ for all steps\;

\textbf{(2) Critic update (CQL).}\;
Initialize $\phi$, $\bar\phi\leftarrow\phi$\;
\While{not converged}{
Sample $(s,a,r,s')\sim\mathcal{D}$\;
$y \leftarrow r + \gamma\,\mathbb{E}_{a'\sim\mathcal{D}(\cdot|s')}\!\left[Q_{\bar\phi}(s',a')\right]$\;
$\phi \leftarrow \arg\min_{\phi}\;
(Q_\phi(s,a)-y)^2 + \alpha\,\mathcal{R}_{\text{cons}} + \lambda_{\text{fail}}\mathcal{L}_{\text{fail}}$\;
\If{iteration mod $N_{\text{tgt}}=0$}{ $\bar\phi \leftarrow \phi$ }
}

\textbf{(3) Actor update (success-guided).}\;
\While{not converged}{
Sample $s\sim\mathcal{D}$, $a^{\text{tele}}\sim\mathcal{D}(\cdot|s)$\;
$\theta \leftarrow \arg\min_{\theta}\;
Q_\phi\!\big(s,\pi_\theta(s)\big)
+\lambda_{\text{anchor}}\|\pi_\theta(s)-a^{\text{tele}}\|^2$\;
}

\Return $Q_\phi,\pi_\theta$\;
\end{algorithm}

\subsection{Value-Guided Impedance Assistance for Bimanual Teleoperation}

We now describe how the success-guided actor policy is rendered to the human operator through impedance feedback on the leader arms. The system adopts a bilateral teleoperation architecture with human-operated master (leader) arms and robot slave (follower) manipulators. Rather than directly modifying operator commands, assistance is conveyed by modulating the mechanical impedance of the leader arms, allowing guidance to be perceived through physical interaction.

As introduced in the previous section, the actor outputs a success-guided assistance action $a^{\text{assist}}$ that encodes a locally safer direction of motion under the learned feasibility landscape. This action is instantiated in the follower space as an incremental motion $\delta \tilde q_{f,i}$. To convey this guidance to the human operator, the follower-side increment is mapped to a leader-side virtual reference using the inverse joint coupling $S^{-1}$ and saturated by $\delta q_{\max}$ to bound the magnitude of assistance. 

\paragraph{Critic-to-Intensity Mapping.}
In parallel, the critic evaluates the current teleoperation state and outputs a scalar success score
\begin{equation}
Q_t \doteq Q_{\phi}(s_t,a_t^{\text{tele}}),
\end{equation}
which is normalized using Euqation. \ref{eq:normlalize} and mapped to a bounded guidance intensity $g_t \in [0,1]$, where lower predicted feasibility corresponds to stronger assistance:
\begin{equation}
g_t
=
\mathrm{clip}\!\left(
\sigma\!\left(\kappa(\tau_g-\bar{Q}_t)\right),\,0,\,1
\right).
\label{eq:g_from_q}
\end{equation}
The intensity signal is low-pass filtered at the servo rate to ensure smooth and stable haptic feedback.

\paragraph{Value-Guided Impedance Rendering.}
Guidance is rendered on each leader arm as a joint-space velocity-attractor impedance,
\begin{equation}
\tau^{\text{guide}}_{\ell,i}
=
K_v(g)\big(\dot q^{\text{des}}_{\ell,i}-\dot q_{\ell,i}\big)
-
D_0\,\dot q_{\ell,i},
\qquad i\in\{L,R\},
\label{eq:impedance}
\end{equation}
where $\dot q^{\text{des}}_{\ell,i}$ denotes the leader-side virtual velocity reference induced by the actor, and $K_v(g)$ is a positive semidefinite diagonal gain modulated by the critic-derived intensity,
\begin{equation}
K_v(g)=K_{\min}+g\,(K_{\max}-K_{\min}),
\quad 0\preceq K_{\min}\preceq K_{\max}.
\end{equation}
$D_0$ is a constant damping matrix. All guidance torques are subject to conservative magnitude and rate limits to ensure safe physical human--robot interaction.

\paragraph{Impedance-mediated intervention.}
This formulation decouples the \emph{direction} of assistance, determined by the actor through the virtual velocity reference $\dot q^{\text{des}}_{\ell,i}$, from the \emph{strength} of assistance, regulated by the critic through $g$. This velocity serves as a virtual motion attractor in the leader-side impedance controller, providing directional guidance that is felt through impedance feedback while preserving the operator’s direct control of motion. Rather than injecting explicit corrective forces or replacing operator commands, the impedance controller makes motions aligned with failure-prone directions increasingly resistive, while preserving compliance along safer directions. 

As a result, the human operator experiences guidance as a gradual increase in resistance or damping near unsafe regions. Through bilateral coupling, this impedance shaping influences execution via physical interaction, enabling proactive risk-aware assistance while preserving continuous and intuitive human control.

\LinesNotNumbered
\begin{algorithm}[htbp]
\caption{Value-Guided Impedance Assistance for Bimanual Teleoperation}
\label{alg:value_guided_impedance}

\KwIn{Joint coupling $S$, policy period $\Delta t_{\pi}$, servo period $\Delta t_s$,
actor $\pi_\theta$, critic $Q_\phi$}
\KwIn{Bounds $(Q_{\min},Q_{\max})$, sigmoid $(\kappa,\tau_g)$,
impedance $(K_{\min},K_{\max},D_0)$, limits $(\delta q_{\max},\tau_{\max})$}

Initialize $g\!\leftarrow\!0$, $\dot q^{\text{des}}_{\ell,i}\!\leftarrow\!\mathbf{0}$, $i\!\in\!\{L,R\}$\;

\While{teleoperation active}{
\If{policy update available ($\Delta t_{\pi}$)}{
$a^{\text{assist}}\!\leftarrow\!\pi_\theta(s_t)$,\;
$Q_t\!\leftarrow\!Q_\phi(s_t,a_t^{\text{tele}})$\;

Actor output $\rightarrow$ follower increment $\delta\tilde q_{f,i}$\;

$\delta q^{\text{ref}}_{\ell,i}\!\leftarrow\!
\mathrm{sat}(S^{-1}\delta\tilde q_{f,i},\delta q_{\max})$\;

$\dot q^{\text{des}}_{\ell,i}\!\leftarrow\!
\delta q^{\text{ref}}_{\ell,i}/\Delta t_{\pi}$\;

$\tilde Q_t\!\leftarrow\!
\mathrm{clip}\!\big((Q_t\!-\!Q_{\min})/(Q_{\max}\!-\!Q_{\min}),0,1\big)$\;

$g\!\leftarrow\!
\mathrm{clip}\!\big(\sigma(\kappa(\tau_g-\tilde Q_t)),0,1\big)$\;
}

(Optional) low-pass filter $g$ at $\Delta t_s$\;

$K_v\!\leftarrow\!K_{\min}+g(K_{\max}-K_{\min})$\;

$\tau^{\text{guide}}_{\ell,i}\!\leftarrow\!
K_v(\dot q^{\text{des}}_{\ell,i}-\dot q_{\ell,i})-D_0\dot q_{\ell,i}$\;

$\tau^{\text{guide}}_{\ell,i}\!\leftarrow\!
\mathrm{sat}(\tau^{\text{guide}}_{\ell,i},\tau_{\max})$\;

Apply guidance torque to leader controller\;
}
\end{algorithm}

\begin{figure}[t]
\centering
\includegraphics[width=0.99\linewidth]{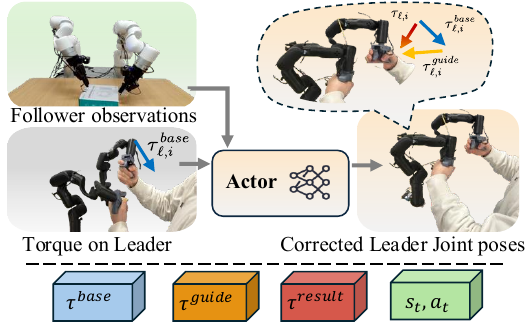}
\caption{\textbf{Value-Guided Impedance Assistance.} Activated guidance generates compliant leader-side torques that discourage unsafe deviations while preserving continuous human authority.}
\label{fig:impedance}
\vspace{-0.5cm}
\end{figure}

\section{Experiments}

\subsection{Experimental Setup}
\subsubsection{Task Design}
\begin{figure*}[htbp]
  \centering
  \includegraphics[width=0.99\linewidth]{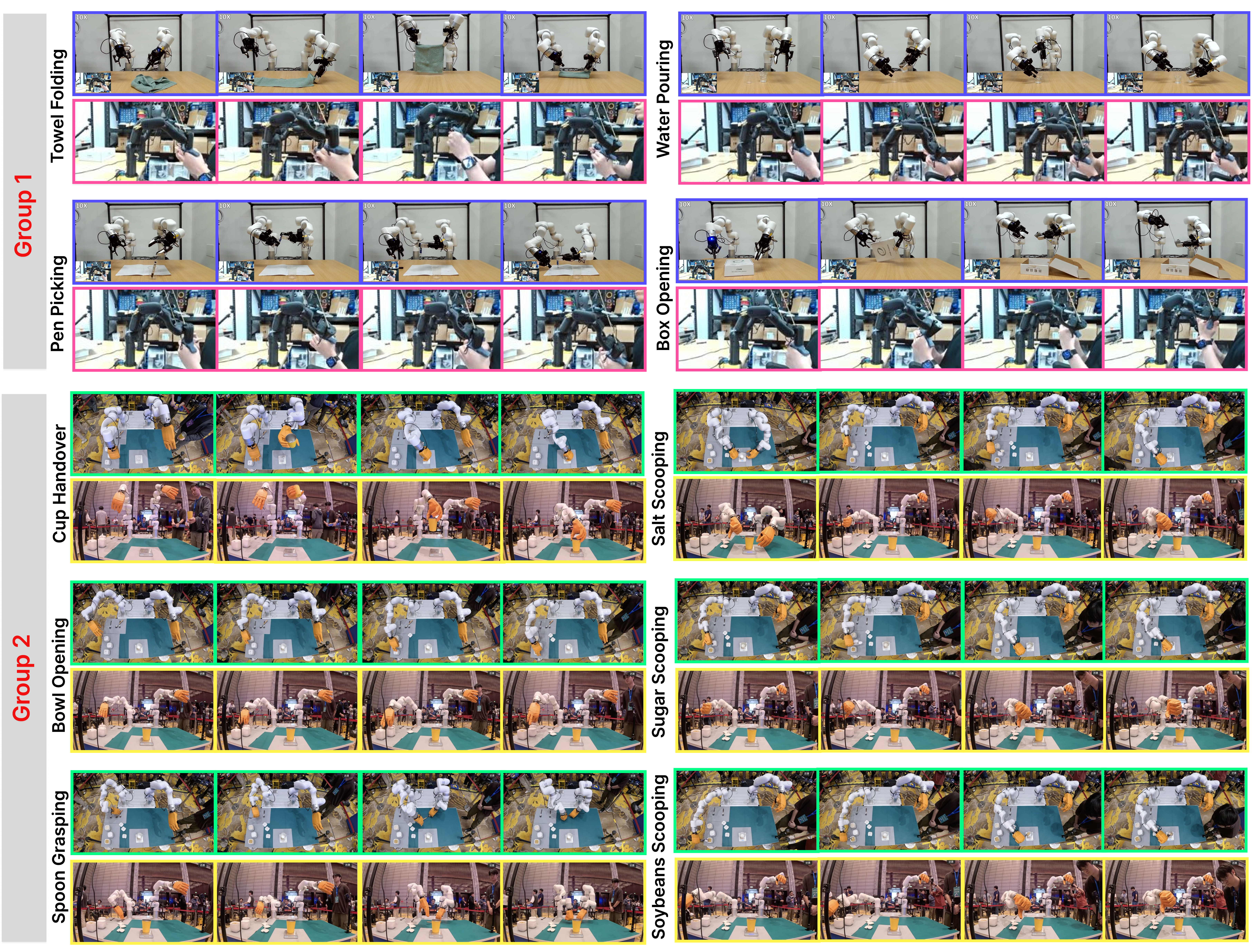}
  \caption{The snapshots of all ten experiments, including four tasks in Group 1 (basic manipulation tasks) with follower-view (in purple box) and leader-view (in pink box), as well as six tasks in Group 2 (interactive fine-grained manipulation tasks) with follower-view (in yellow box) and top-view (in green box).}
  \label{fig:experiment-1}
\end{figure*}

Ten representative daily-life robotic manipulation tasks are designed to comprehensively verify the adaptability of our methods across diverse operation paradigms. According to the operation complexity and interaction attributes, these tasks are divided into two groups: Group 1 (Basic Manipulation Tasks) involves relatively simple, single-stage operations with no or minimal inter-object interaction; Group 2 (Interactive Fine-grained Manipulation Tasks) requires sequential multi-step execution, inter-object interaction, or precise control of small objects/quantities. We use the teleoperation system introduced in our previous work, Unibidex \cite{li2025unibidex}. The snapshots of these experiments are shown in Fig.~\ref{fig:experiment-1}, respectively.

\textbf{Group 1} includes four tasks: each completed via a core single operation: (1) Towel Folding: Fold a rectangular cotton towel (20cm×45cm) into a 10cm×10cm small rectangle and place it at a preset target position. This task focuses on soft object shaping with a single folding action. (2) Pen Picking: Take the marker pen out of the pen holder on the desktop, then pull off the pen cap, and finally place the marker pen at the designated position on the paper. The core requirements are rigid-object grasping and vertical lifting. (3) Water Pouring: Pour 100ml of water from a cylindrical cup (capacity 200ml, diameter 8cm) into another cylindrical cup without spilling, which focuses on liquid transfer via a single pouring motion. (4) Box Opening: Open a flip-top cardboard box (15cm×10cm×8cm) by lifting the lid to an angle of $>120^\circ$, then take out the tape measure from the box and use it to measure the length of the box opening.

\textbf{Group 2} consists of six tasks that demand multi-step coordination or precise interaction: (5) Cup Handover: A robot arm picks up a stainless steel cup from its initial position on the table and smoothly transfers it to another robot arm located more than 60 centimeters away from the initial position, where this robot arm places the stainless steel cup at a specified location on the table. (6) Bowl Opening: A robot arm needs to sequentially remove the lids of the three ceramic bowls that are randomly placed on the table. (7) Spoon Grasping: A robot arm picks up a stainless steel spoon from its initial position on the table. (8)-(10) Salt/Rice/Soybeans Scooping: The robot arm that grabs the spoon (in Task (7)) scoops up the granular materials from the three open ceramic bowls (from Task (6)) and places them in the stainless steel cup on the table (from Task (5)). Three ceramic bowls were filled with salt, rice, and soybeans, respectively.

\begin{table*}[htbp]
\centering
\caption{Completion time and success rate of different methods on all ten tasks.}
\label{tab:result_all_ten_taks}
\resizebox{\linewidth}{!}{
  \begin{tabular}{@{}c*{20}{c}@{}} 
    \toprule
    & \multicolumn{8}{c}{\textbf{\normalsize Group 1}} & \multicolumn{12}{c}{\textbf{\normalsize Group 2}} \\ 
    \cmidrule(lr){2-9} \cmidrule(lr){10-21}
    \multirow{2}{*}{\textbf{\normalsize Method}} & \multicolumn{2}{c}{\shortstack{\textbf{\normalsize Towel}\\\textbf{\normalsize Folding}}} & \multicolumn{2}{c}{\shortstack{\textbf{\normalsize Pen}\\\textbf{\normalsize Picking}}} & \multicolumn{2}{c}{\shortstack{\textbf{\normalsize Water}\\\textbf{\normalsize Pouring}}} & \multicolumn{2}{c}{\shortstack{\textbf{\normalsize Box}\\\textbf{\normalsize Opening}}} & \multicolumn{2}{c}{\shortstack{\textbf{\normalsize Cup}\\\textbf{\normalsize Handover}}} & \multicolumn{2}{c}{\shortstack{\textbf{\normalsize Bowl}\\\textbf{\normalsize Opening}}} & \multicolumn{2}{c}{\shortstack{\textbf{\normalsize Spoon}\\\textbf{\normalsize Grasping}}} & \multicolumn{2}{c}{\shortstack{\textbf{\normalsize Salt}\\\textbf{\normalsize Scooping}}} & \multicolumn{2}{c}{\shortstack{\textbf{\normalsize Rice}\\\textbf{\normalsize Scooping}}} & \multicolumn{2}{c}{\shortstack{\textbf{\normalsize Soybeans}\\\textbf{\normalsize Scooping}}} \\
    \cmidrule(lr){2-3} \cmidrule(lr){4-5} \cmidrule(lr){6-7} \cmidrule(lr){8-9} \cmidrule(lr){10-11} \cmidrule(lr){12-13} \cmidrule(lr){14-15} \cmidrule(lr){16-17} \cmidrule(lr){18-19} \cmidrule(lr){20-21}
    & Time(s) & Succ & Time(s) & Succ & Time(s) & Succ & Time(s) & Succ & Time(s) & Succ & Time(s) & Succ & Time(s) & Succ & Time(s) & Succ & Time(s) & Succ & Time(s) & Succ \\
    \midrule
    \normalsize
    Human    & $22\pm5$  & 40/40 & $7\pm2$  & 40/40 & $12\pm3$  & 40/40 & $23\pm4$  & 40/40 & $8\pm1$  & 40/40 & $14\pm3$ & 40/40 & $19\pm3$ & 40/40 & $9\pm2$  & 40/40 & $11\pm3$ & 39/40 & $12\pm3$ & 38/40 \\
    \normalsize FACTR    & $198\pm6$ & 35/40 & $36\pm4$ & 36/40 & $87\pm8$  & 37/40 & $163\pm6$ & 34/40 & $77\pm8$ & 38/40 & $89\pm9$ & 36/40 & $192\pm13$& 34/40 & $46\pm8$ & 37/40 & $48\pm8$ & 37/40 & $54\pm9$ & 36/40 \\
    \normalsize GELLO    & $179\pm12$& 32/40 & $42\pm5$ & 32/40 & $93\pm12$ & 35/40 & $176\pm10$& 31/40 & $89\pm9$ & 36/40 & $108\pm12$& 33/40 & $231\pm18$& 31/40 & $58\pm12$& 34/40 & $52\pm9$ & 33/40 & $62\pm20$& 31/40 \\
    \normalsize VR       & $234\pm27$& 31/40 & $52\pm9$ & 33/40 & $104\pm17$& 34/40 & $252\pm26$& 30/40 & $98\pm11$& 34/40 & $186\pm17$& 32/40 & $328\pm34$& 28/40 & $67\pm14$& 31/40 & $74\pm11$& 29/40 & $70\pm12$& 27/40 \\
    \normalsize Ours-LF  & $157\pm6$ & 39/40 & $31\pm4$ & 40/40 & $78\pm8$  & 40/40 & $144\pm5$ & 39/40 & $60\pm7$  & 40/40 & $78\pm8$  & 40/40 & $174\pm13$& 38/40 & $37\pm4$  & 40/40 & $41\pm4$  & 39/40 & $44\pm5$  & 39/40 \\
    \bottomrule
  \end{tabular}
}
\end{table*}

\begin{figure*}[htbp]
  \centering
  \includegraphics[width=0.99\linewidth]{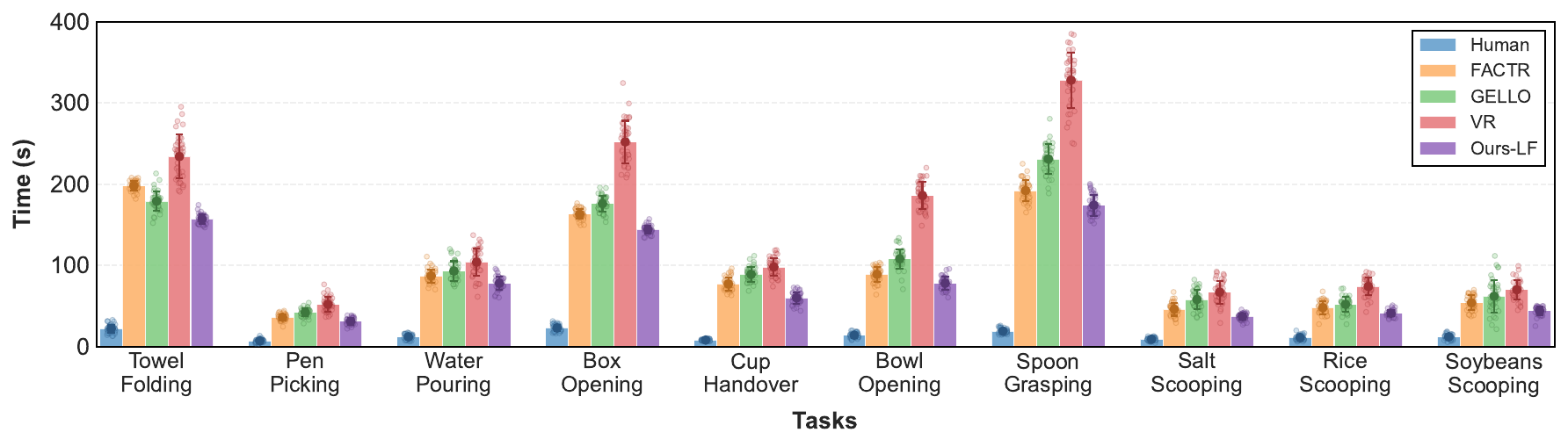}
  \caption{Comparison of completion time for all tasks under baselines and our method.}
  \label{fig:result_completion_time}
  \vspace{-1cm}
\end{figure*}

\subsubsection{Platforms}
The experimental platform is built around a leader--follower teleoperation architecture comprising a user-operated leader arm and two remotely controlled follower arms, as illustrated in Fig. \ref{fig:method-1}. This setup is specifically designed to facilitate the collection of high-fidelity manipulation data, as described in Sec. \ref{method:collection}.
The follower arms are two identical 7-DoF collaborative robotic arms (UFACTORY xArm 7), each with a maximum payload of 3.5 kg. These arms provide the necessary dexterity and precision for executing diverse manipulation tasks. The master arms are two custom-designed, 3D-printed manipulators that mirror the kinematic structure of the xArm 7. This design ensures an intuitive one-to-one mapping between the operator's hand movements and the follower arm's motions, minimizing the cognitive load during teleoperation. In addition, a foot pedal is placed beneath the data collection table and used by the operator to confirm actions.
To match the varying requirements of the two task groups, different end-effectors are employed. For Group 1 tasks, a two-finger adaptive gripper (Robotiq 2F-85) is used. This gripper offers a maximum stroke of 85mm and a grasping force range of 10-140N, making it suitable for handling objects of various sizes and stiffnesses, from soft towels to rigid pens and boxes. For Group 2 tasks, a dexterous robotic hand (DH-5-6) is utilized. This 5-fingered hand provides 16 DoF in total, enabling complex grasping and manipulation strategies required for tasks such as spoon handling, precise pouring, and object handover. Its fine-grained control is essential for the delicate operations involving small objects and granular materials.

\subsection{Comparison Methods}
We compare our method against four baselines. Specifically, \textbf{Human} denotes direct manual execution by a person using their hands in the physical environment, without any robot teleoperation, serving as a reference for human-level performance with rich tactile feedback. \textbf{GELLO}~\cite{wuphilipp2024gello} is a kinematically matched teleoperation interface implemented as a leader--follower controller without learned assistance. \textbf{VR}~\cite{ding2025bunny} uses a virtual-reality-based bimanual teleoperation interface whose commands are tracked by the robot in a leader--follower manner, also without learned assistance. \textbf{FACTR}~\cite{liu2025factr} is a low-cost \emph{bilateral} leader--follower teleoperation system that actuates the leader arm to relay the follower’s external joint torques, providing force feedback (with additional gravity compensation and redundancy resolution), thereby improving teleoperation performance in contact-rich tasks. \textbf{Ours-LF} is our method, where a conservative success score learned from offline teleoperation data modulates a continuous guidance intensity, and a learned actor provides a preferred corrective direction; both are rendered as compliant joint-space impedance haptic guidance to bias the operator away from failure-prone actions while preserving human authority.

\subsection{Implementation}
To ensure the reliability and statistical significance of the experimental results, the following protocol is strictly implemented: (1) Trial Repetition: Each task is repeated 40 times for each of the five methods (Ours-LF, Human, GELLO, VR, Naive-LF). The 40 repetitions for each task-method combination are conducted consecutively to avoid environmental changes, and a 2-minute interval is set between different task-method combinations to prevent equipment overheating and operator fatigue (for Human and VR methods). (2) Operator Selection for Human/VR Methods: Three experienced operators (with more than 3 months of robotic teleoperation experience) are recruited to perform the tasks using the Human and VR methods. All operators receive a 1-hour training session before the formal experiments to familiarize themselves with the task requirements and operation tools. (3) Data Initialization: For Ours-LF and Naive-LF, the teleoperation data collected and annotated in Sec. \ref{method:collection} (including 500 valid operation trajectories for the 10 tasks) are used to train the model. The training process is conducted on a server equipped with an Intel Core i9-12900K CPU and an NVIDIA RTX 5090 GPU, and the model parameters are fixed after 100 epochs to ensure consistent initial conditions across all trials. (4) Data Recording: During each trial, the following data are recorded in real time: the robotic arm’s joint angles, end effector position/orientation, gripper opening/closing degree, left/right pedal actions, and the start/end time of the task.

\subsection{Experimental Results}
Two key metrics are used to evaluate the experiment results of all ten tasks. \textbf{Completion Time}: Time taken from task start to successful completion. Failures are not recorded in the completion time. \textbf{Success Rate}: Percentage of trials completed successfully within the time limit.
The statistical results of completion time and success rate for the 10 tasks under the five comparative methods are presented in Table \ref{tab:result_all_ten_taks}, respectively. In detail, the completion times are further illustrated in Fig. \ref{fig:result_completion_time}, which intuitively reflects the efficiency differences among the five methods in task execution.

For Group 1 tasks, the Human operator serves as the performance upper bound, achieving the shortest completion times (e.g., $7\pm2$ s for Pen Picking) and a $100\%$ success rate (40/40) across all four tasks. Among teleoperation methods, Ours-LF demonstrates superior efficiency: for Towel Folding, its completion time ($157\pm6$ s) is notably shorter than FACTR ($198\pm6$ s), GELLO ($179\pm12$ s), and VR ($234\pm27$ s), paired with a near-perfect success rate (39/40). For Pen Picking, Water Pouring, and Box Opening, Ours-LF outperforms other teleoperation methods in completion time ($31\pm4$ s, $78\pm8$ s, $144\pm5$ s, respectively) while maintaining high success rates (40/40 for Pen Picking and Water Pouring, 39/40 for Box Opening), resulting in a performance comparable to the Human baseline.

For Group 2 tasks, the Human operator serves as the performance baseline, exhibiting the shortest completion times across most tasks (e.g., $19\pm3$ s for Spoon Grasping) and near-perfect success rates (40/40 for four tasks, with 39/40 for Rice Scooping and 38/40 for Soybeans Scooping). Among teleoperation approaches, Ours-LF demonstrates superior efficiency and reliability. In Cup Handover, its completion time ($60\pm7$ s) is shorter than FACTR ($77\pm8$ s), GELLO ($89\pm9$ s), and VR ($98\pm11$ s), with a $100\%$ success rate (40/40). For Bowl Opening, Ours-LF reduces completion time to $78\pm8$ s (vs. over $89\pm9$ s for other baselines) while achieving $100\%$ success. In fine-grained tasks like Salt Scooping, Ours-LF achieves the shortest completion time ($37\pm4$ s) and $100\%$ success rate, outperforming FACTR ($46\pm8$ s), Gello ($58\pm12$ s), and VR ($67\pm14$ s). For Spoon Grasping, Rice Scooping, and Soybeans Scooping, Ours-LF achieves shorter completion times and higher success rates ($\geqslant$38/40) than other teleoperation methods.



\begin{figure}[htbp]
  \centering
  \includegraphics[width=0.99\linewidth]{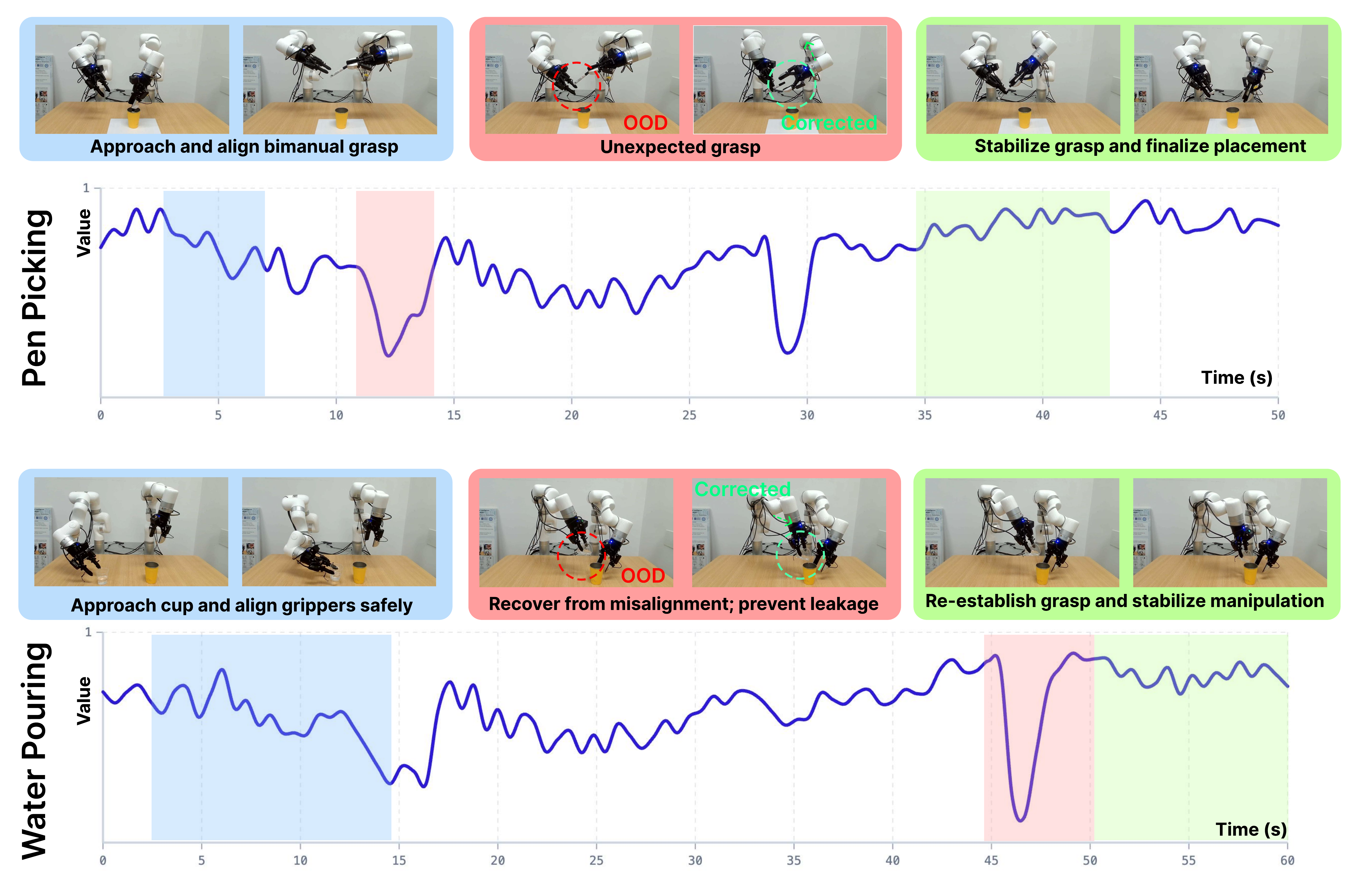}
  \caption{Failure correction. During remote operation, our failure-aware framework detects OOD conditions and intervenes actively, promptly applying appropriate actions to the leader arms, thereby enabling the follower arms to return to their normal states (shown in green dashed arrows).}
  \label{fig:exp_ood}
  \vspace{-0.5cm}
\end{figure}

\subsection{Result Analysis}
From above experimental results, it can be found that the Human baseline achieves a $\sim$100$\%$ average success rate and the shortest completion times across all tasks. For Ours-LF, its average success rate reaches $\geqslant$98$\%$ (vs. $\sim$80$\%$ for FACTR/GELLO/VR), and its completion time is $\sim$25$\%$ shorter than other teleoperation methods on average, as seen in Fig. \ref{fig:result_completion_time}.
The superior performance of Ours-LF stems from two core innovations in the proposed teleoperation framework that address key limitations of existing approaches.

First, our framework enables robust identification of OOD conditions and real-time correction, i.e., the failure-aware correction mechanism. Upon detecting OOD states (e.g., the pen cap outside the range of the gripper shown in Fig. \ref{fig:exp_ood}), the framework issues targeted motor commands to the leader arms. This translates to the follower robot arms adjusting their trajectories to revert to normal operational ranges. This mechanism reduces the time required to recover from deviations while alleviating operator cognitive load, allowing users to focus on core manipulation tasks rather than error recovery directly. This contributes to Ours-LF’s shorter completion times and near-perfect success rates.

Second, integrated force-control feedback in our framework enhances performance especially in contact-rich manipulation tasks. For example, in the task of Spoon Grasping, environmental forces on the follower robot arms are rapidly relayed to the leader arms, enabling adaptive action modulation. Methods lacking force feedback (e.g., VR, GELLO) frequently trigger emergency stops due to unregulated robot-environment impacts, resulting in the lowest success rates in all tasks in Group 2 (see Table \ref{tab:result_all_ten_taks}). While FACTR incorporates force control, its lack of failure-aware correction leads to longer completion times and lower success rates compared to Ours-LF. In summary, the synergy of failure-aware correction and force feedback underpins Ours-LF’s superior performance across all 10 tasks.


\subsection{Limitations}
Despite promising results, the proposed framework has several limitations. First, the guidance actor optimizes corrective directions based on the learned success score but does not explicitly enforce contact constraints, such as unilateral contact or frictional modes. Consequently, the suggested motions may conflict with instantaneous contact states, leading to local oscillations and inconsistent haptic feedback in contact-rich interactions. Second, failure awareness is inherently limited by the support of the offline dataset. Although the conservative success score improves robustness to operator-induced distribution shift, its reliability may degrade under changes in task conditions (e.g., end effector geometry, materials), necessitating additional data or adaptation to maintain consistent guidance.

\section{Conclusion}

This paper presented a failure-aware shared-autonomy framework for bimanual teleoperation in contact-rich manipulation. Using heterogeneous offline teleoperation data with both successes and failures, we learned a conservative success score that provides risk-sensitive feasibility estimates under distribution shift. During execution, this score modulates assistance strength, while a learned actor supplies corrective motion directions, both rendered through a joint-space impedance interface to deliver compliant, physically grounded haptic guidance without compromising human authority. Experiments on ten daily-life manipulation tasks show consistent improvements in success rate and efficiency over conventional teleoperation and shared-autonomy baselines. Overall, the results indicate that conservative value learning provides an effective means to embed predictive failure awareness into bilateral teleoperation, improving reliability in contact-rich settings.

\bibliographystyle{IEEEtran}
\bibliography{refs}

\vspace{-1.5cm}
\begin{IEEEbiography}
[{\includegraphics[width=1in,height=1.25in,clip,keepaspectratio]{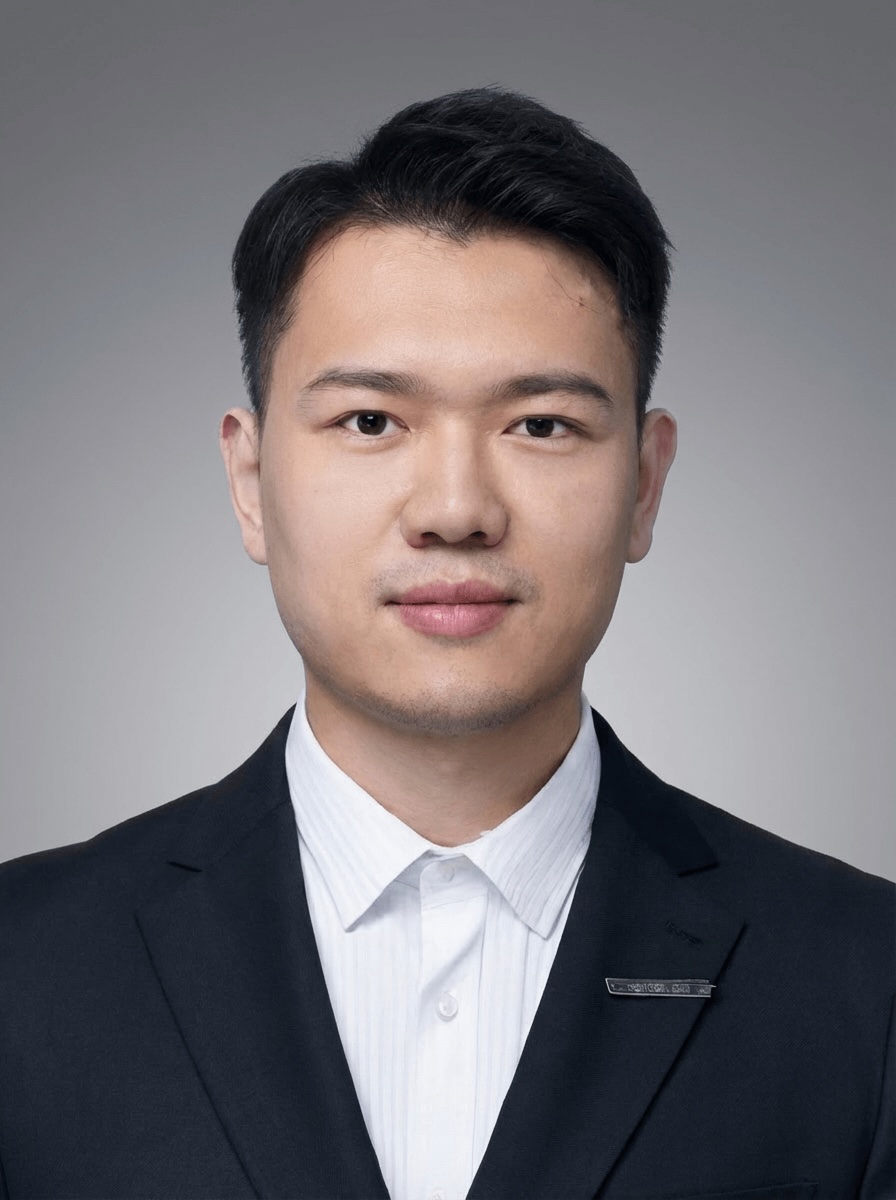}}] 
{Peng Zhou} received his Ph.D. degree in robotics from The Hong Kong Polytechnic University, Hong Kong SAR, in 2022. In 2021, he was a Visiting Researcher at the Robotics, Perception, and Learning Lab, KTH Royal Institute of Technology, Stockholm, Sweden.
From 2022 to 2024, he was a Postdoctoral Research Fellow at the Department of Computer Science, The University of Hong Kong. He is currently an Assistant Professor in the School of Advanced Engineering, Great Bay University (GBU).
His research interests include robot manipulation, robot learning, and task and motion planning. Dr. Zhou currently serves as an Associate Editor of the \textsc{IEEE Robotics and Automation Letters}
and \textsc{Frontiers in Robotics and AI}.
\end{IEEEbiography}
\vspace{-1cm}

\begin{IEEEbiography}
[{\includegraphics[width=1in,height=1.25in,clip,keepaspectratio]{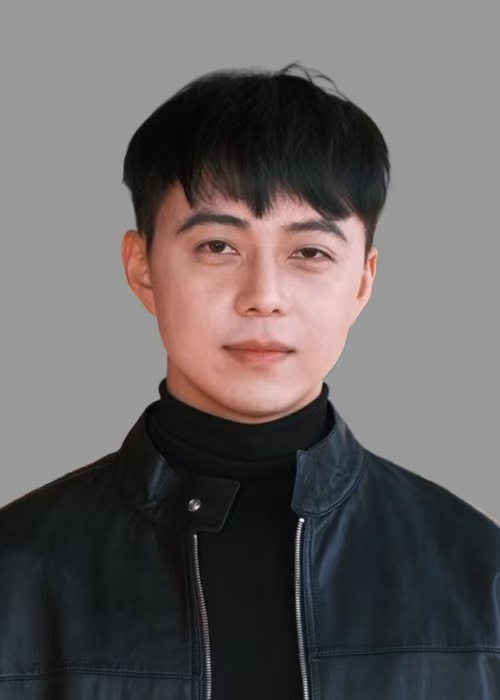}}] 
{Zhongxuan Li} received the integrated M.Eng. degree (with the Bachelor's degree incorporated) in Electrical and Electronic Engineering from Imperial College London, London, U.K.
He is currently pursuing the Ph.D. degree in Robotics with the Department of Computer Science, The University of Hong Kong (HKU), Hong Kong. He is also a visiting student with the Email Lab, Great Bay University. 
His research interests include robot task understanding and planning.
\end{IEEEbiography}
\vspace{-0.5cm}

\begin{IEEEbiography}
[{\includegraphics[width=1in,height=1.25in,clip,keepaspectratio]{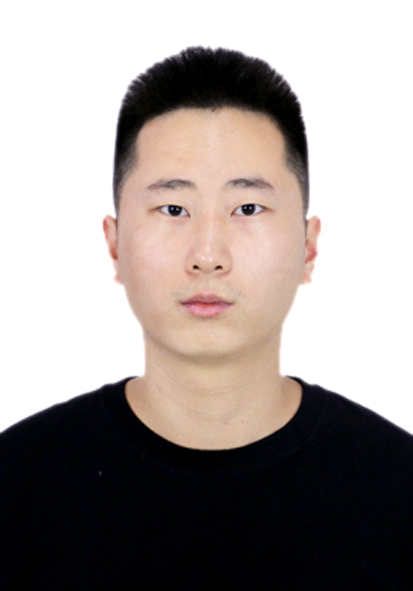}}]
{Jinsong Wu} received the BEng degree in mechatronics and robotic systems from the University of Liverpool, UK, in 2021, and the MRes degree in medical robotics and image-guided intervention from Imperial College London, UK, in 2022. 
From 2022 to 2023, he was a Research Assistant the The Chinese University of Hong Kong, Hong Kong.
Since 2023, he has been pursuing a Ph.D. degree in mechanical engineering at The Hong Kong Polytechnic University, Hong Kong. His current research interests include medical robotics and control theory.
\end{IEEEbiography}
\vspace{-0.5cm}

\begin{IEEEbiography}
[{\includegraphics[width=1in,height=1.25in,clip,keepaspectratio]{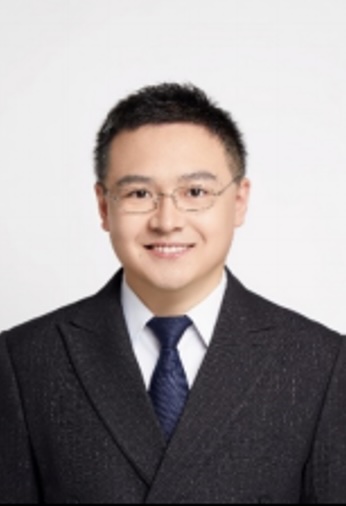}}]
{Jiaming Qi} received the Ph.D. degree in Control Science and Engineering and the M.S. degree in Integrated Circuit Engineering from Harbin Institute of Technology, Harbin, China, in 2023 and 2018, respectively. 
In 2019, he was a visiting Ph.D. student at the Robotics and Machine Intelligence Laboratory, The Hong Kong Polytechnic University. 
In 2023, he served as a Postdoctoral Research Fellow in the Centre for Garment Production Limited, Department of Computer Science, University of Hong Kong, Hong Kong SAR, China.
He is currently an Assistant Professor in the College of Mechanical and Electrical Engineering at Northeast Forestry University, Harbin, China.
His research interests lie in the areas of deformable object manipulation, visual servoing, human-robot collaboration, and robot compliance control.
\end{IEEEbiography}
\vspace{-0.5cm}

\begin{IEEEbiography}
[{\includegraphics[width=1in,height=1.25in,clip,keepaspectratio]{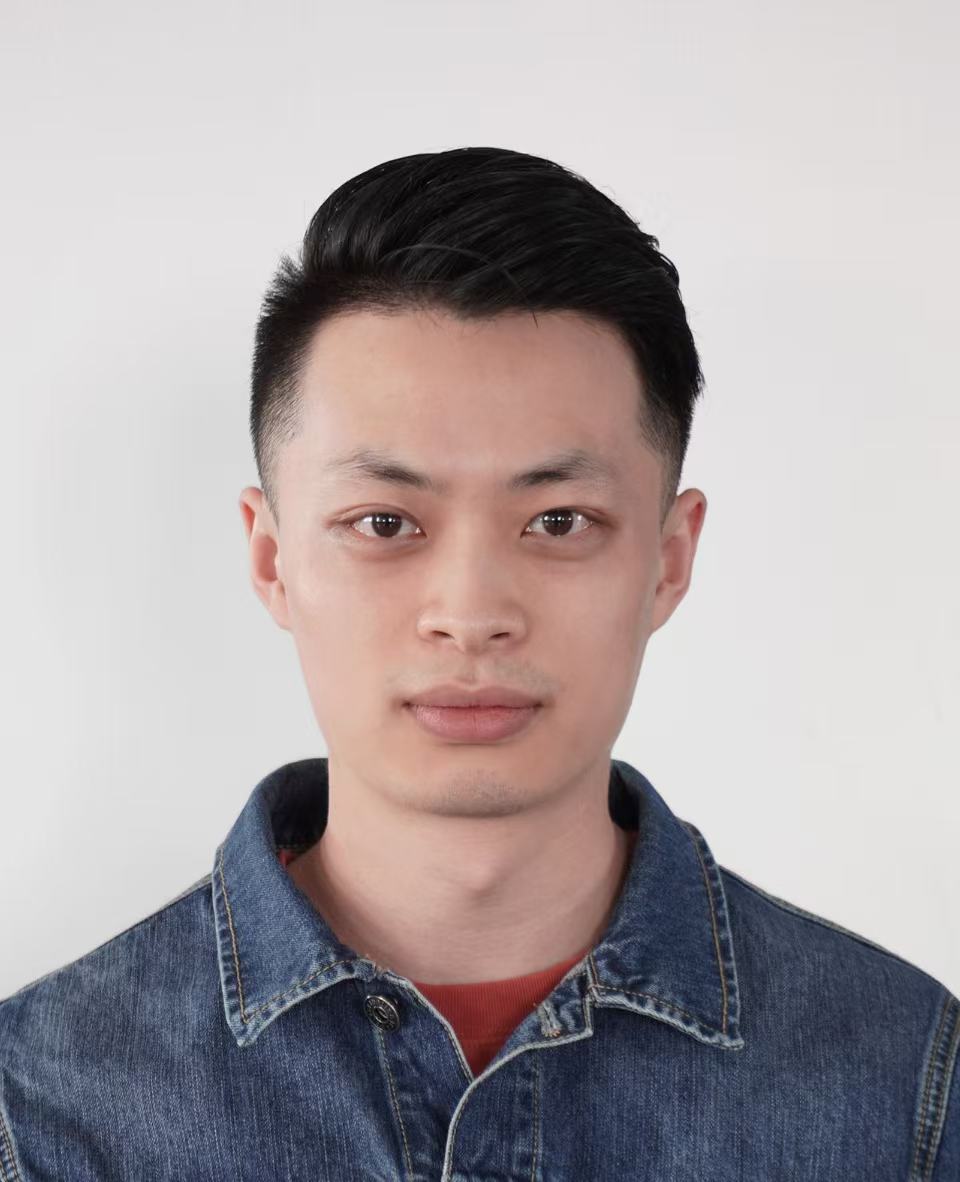}}]
{Jun Hu} received  the Ph.D. degree in control science and engineering from the University of Science and Technology of China, Hefei, China, in 2023.
He is currently a Postdoctoral Researcher with the School of Advanced Engineering, Great Bay University, Guangdong, China. His research interests lie in robotic manipulation and perception, with a particular focus on dexterous manipulation, tactile sensing, multimodal perception, and skill learning for robotic systems. His recent work includes tactile sensor design and calibration, teleoperation for dexterous hands, and learning-based methods for force and contact understanding in robotic manipulation.
\end{IEEEbiography}
\vspace{-0.5cm}

\begin{IEEEbiography}
[{\includegraphics[width=1in,height=1.25in,clip,keepaspectratio]{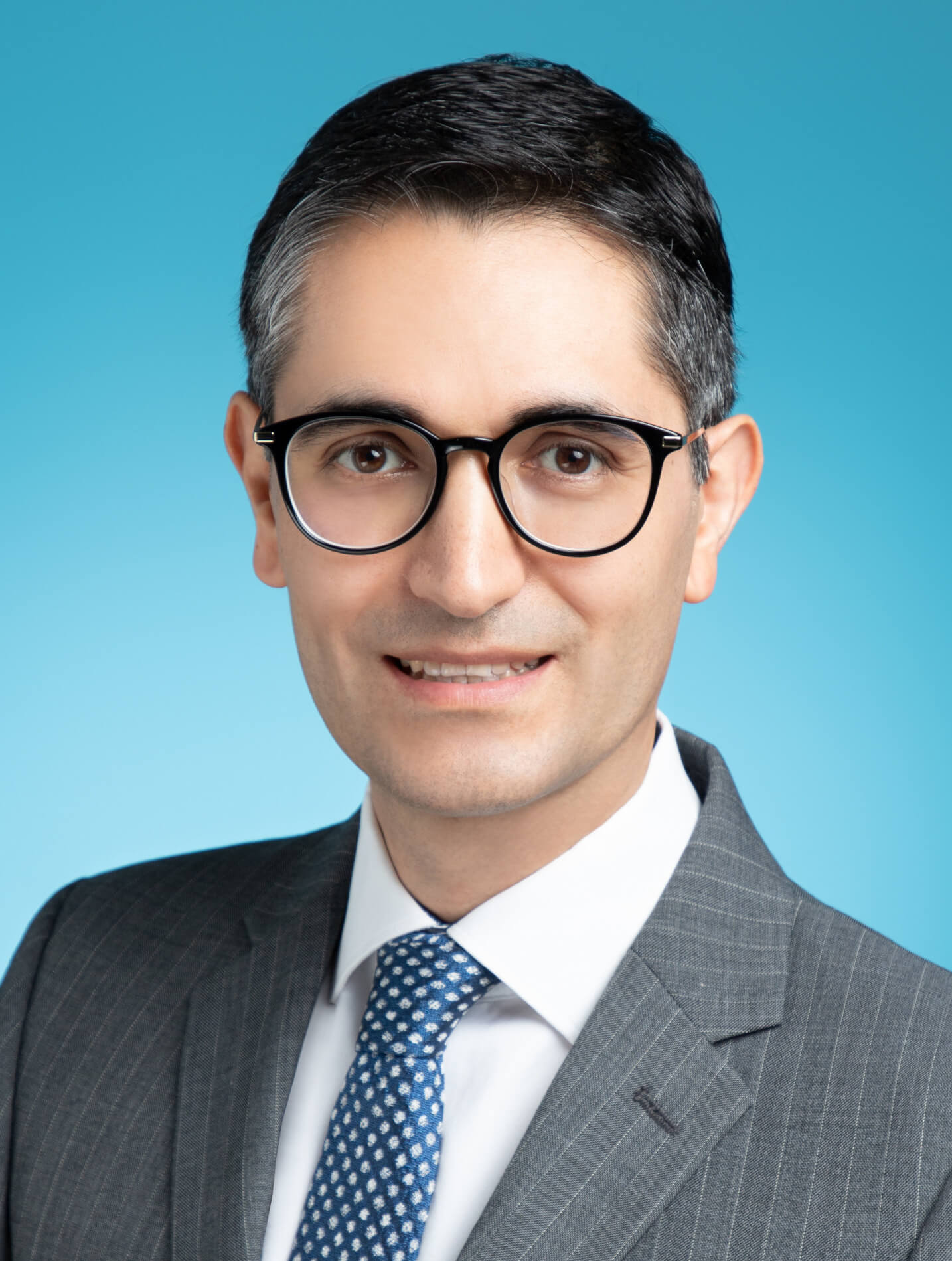}}]
{David Navarro-Alarcon} (Senior Member, IEEE) received the Ph.D. degree in mechanical and automation engineering from The Chinese University of Hong Kong, in 2014. 
Since 2017, he has been with The Hong Kong Polytechnic University, where he is currently an Associate Professor with the Department of Mechanical Engineering.
His current research interests include perceptual robotics and control systems.
He currently serves as an Associate Editor of the \textsc{IEEE Transactions on Robotics}.
\end{IEEEbiography}
\vspace{-0.5cm}

\begin{IEEEbiography}
[{\includegraphics[width=1in,height=1.25in,clip,keepaspectratio]{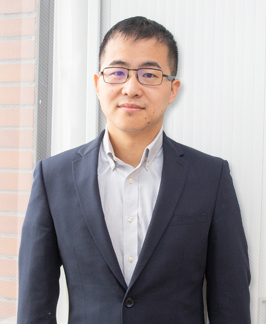}}] 
{Jia Pan} (Senior Member, IEEE) 
received the Ph.D. degree in computer science from the University of North Carolina at Chapel Hill, Chapel Hill, NC, USA, in 2013.
He is currently an Associate Professor with the Department of Computer Science, University of Hong Kong, Hong Kong. He is also a member of the Centre for Garment Production Limited, Hong Kong. His research interests include robotics and artificial intelligence as applied to autonomous systems, particularly for navigation and manipulation in challenging tasks such as effective movement in dense human crowds and manipulating deformable objects for garment automation.
\end{IEEEbiography}
\vspace{-0.5cm}

\begin{IEEEbiography}
[{\includegraphics[width=1in,height=1.25in,clip,keepaspectratio]{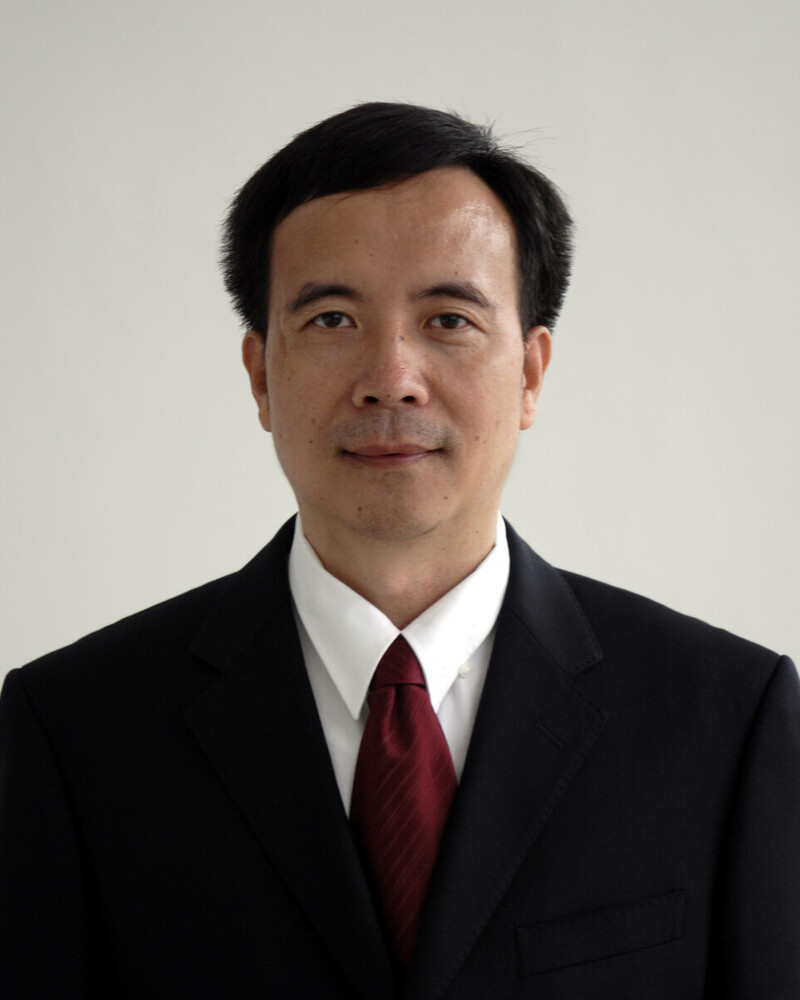}}] 
{Lihua Xie} (Fellow, IEEE) received the B.E. and M.E. degrees in electrical engineering from the Nanjing University of Science and Technology, Nanjing, China, in 1983 and 1986, respectively, and the Ph.D. degree in electrical engineering from the University of Newcastle, Australia, Callaghan, NSW, Australia, in 1992. Since 1992, he has been with the School of Electrical and Electronic Engineering, Nanyang Technological University, Singapore, where he is currently a Professor and the Director of the Center for Advanced Robotics Technology Innovation. He held teaching appointments with the Department of Automatic Control, Nanjing University of Science and Technology from 1986 to 1989. He was the Head of the Division of Control and Instrumentation from July 2011 to June 2014 and Co-Director of Delta-NTU Corporate Lab for Cyber–Physical Systems from 2016 to 2021. His research interests include robust control and estimation, networked control systems, multiagent networks, localization, and unmanned systems. Dr. Xie is the Editor-in-Chief for Unmanned Systems and an Associate Editor for IEEE Control System Magazine. He was the Editor of the IET Book Series in Control and Associate Editor of a number of journals, including IEEE TRANSACTIONS ON AUTOMATIC CONTROL, Automatica, IEEE TRANSACTIONS ON CONTROL SYSTEMS TECHNOLOGY, IEEE TRANSACTIONS ON NETWORK CONTROL SYSTEMS, and IEEE TRANSACTIONS ON CIRCUITS AND SYSTEMS-II. He was an IEEE Distinguished Lecturer (January 2012– December 2014). He is a Fellow of the Academy of Engineering Singapore, IEEE, IFAC, and CAA.
\end{IEEEbiography}
\vspace{-0.5cm}

\begin{IEEEbiography}
[{\includegraphics[width=1in,height=1.25in,clip,keepaspectratio]{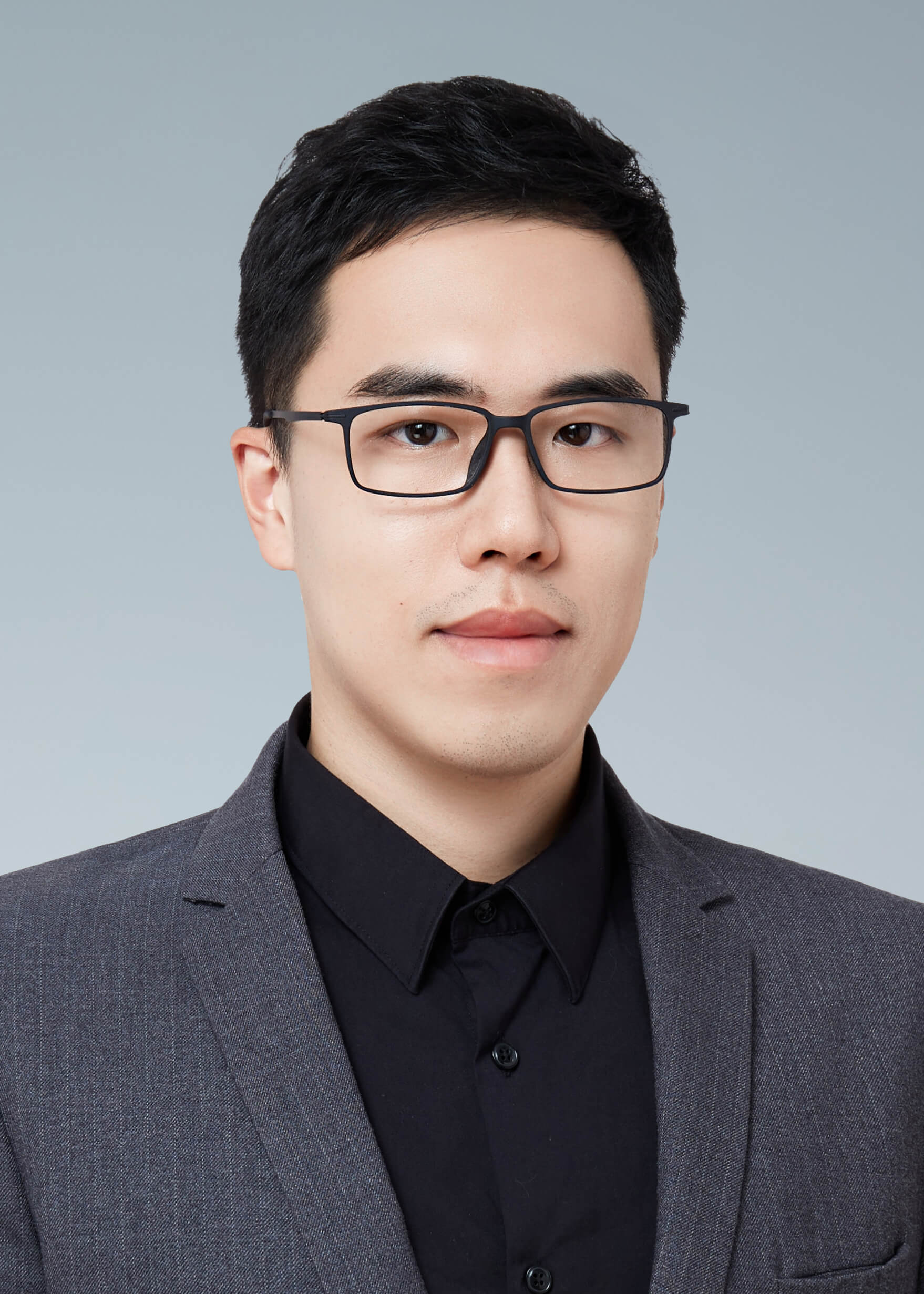}}]
{Shiyao Zhang} (Senior Member, IEEE) is currently an Assistant Professor with the School of Advanced Engineering, Great Bay University. He received the B.S. degree (Hons.) in Electrical and Computer Engineering from Purdue University, West Lafayette, IN, USA, in 2014, the M. S. degree in Electrical Engineering from University of Southern California, Los Angeles, CA, USA, in 2016, and the Ph.D. degree from the University of Hong Kong, Hong Kong SAR, China, in 2020. He was a Post-Doctoral Research Fellow with the Academy for Advanced Interdisciplinary Studies, Southern University of Science and Technology, from 2020 to 2022, and a Research Assistant Professor with the Research Institute for Trustworthy Autonomous Systems, Southern University of Science and Technology, from 2022 to 2024. His research interests include intelligent transportation systems, embodied AI, and transportation electrification. He has published more than 60 academic papers in top international journals and conferences. He is a junior Editor of the \textsc{Intelligence \& Robotics} and \textsc{AI \& Autonomous Systems} journals.
\end{IEEEbiography}
\vspace{-0.5cm}

\begin{IEEEbiography}
[{\includegraphics[width=1in,height=1.25in,clip,keepaspectratio]{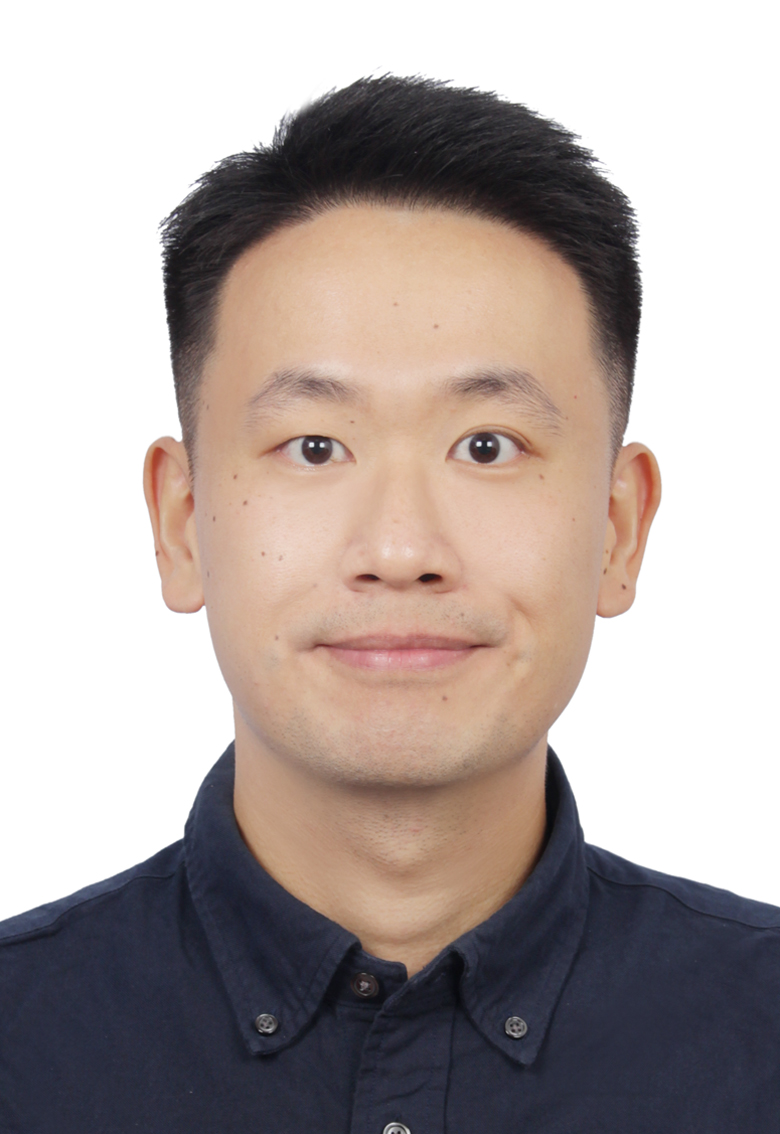}}]
{Zeqing Zhang} (Member, IEEE) received the M.Phil. degree from the Chinese University of Hong Kong in 2019, and the Ph.D. degree in computer science from the University of Hong Kong in 2024. From 2024 to 2025, he served as a Research Associate (Honorary) at the University of Hong Kong and also served as an ITF-funded postdoctoral fellow at InnoHK Centre for Transformative Garment Production (TransGP) in Hong Kong. Now he is a Research Fellow in the Nanyang Technological University in Singapore. His research interests focus on using robotic proprioceptive and exteroceptive sensing for the perception and manipulation of soft bodies (e.g., garments) and granular media (e.g., sands). He primarily works with vision, touch, force sensing, and text modalities to handle these nonlinear materials. 
He as a visiting scholar has visited Osaka University (Japan) in 2024 and Stanford University (United States) in 2025.
\end{IEEEbiography}

\vfill
\end{document}